\documentclass{article}

\usepackage[nonatbib,final]{nips_2018}
\pdfoutput=1
\usepackage[utf8]{inputenc} % allow utf-8 input
\usepackage[T1]{fontenc}    % use 8-bit T1 fonts
\usepackage{url}            % simple URL typesetting
\usepackage{booktabs}       % professional-quality tables
\usepackage{amsfonts}       % blackboard math symbols
\usepackage{nicefrac}       % compact symbols for 1/2, etc.
\usepackage{microtype}      % microtypography
\usepackage{fancyhdr}

\usepackage{graphicx}
\usepackage{subfig}

\usepackage{multirow}
\PassOptionsToPackage{hyphens}{url}\usepackage{hyperref}

\title{Energy Efficient Hardware for On-Device CNN Inference via Transfer Learning}

\author{
  Paul Whatmough\qquad Chuteng Zhou\qquad Patrick Hansen\qquad Matthew Mattina \\
  Arm Research \\ Boston, MA \\
  \texttt{\{Paul.Whatmough,Chu.Zhou,Patrick.Hansen,Matthew.Mattina\}@arm.com}
}

\begin{document}

\maketitle

\begin{abstract}
%  The abstract paragraph should be indented \nicefrac{1}{2}~inch
%  (3~picas) on both the left- and right-hand margins. Use 10~point
%  type, with a vertical spacing (leading) of 11~points.  The word
%  \textbf{Abstract} must be centered, bold, and in point size 12. Two
%  line spaces precede the abstract. The abstract must be limited to
%  one paragraph.

% Mobile devices place a huge demand on the throughput and energy efficiency requirements of computer vision.
% Aggressively optimized hardware accelerators improve performance at the cost of programmability.
% This paper proposes the combination of a heavily optimized fixed-weight feature extractor accelerator that is used in combination with a conventional programmable accelerator.
% CNNs are trained with the common feature extractor layers, which can be generalized to new datasets in analogy to transfer learning.
% This arrangement provides a significant improvement in latency and energy efficiency, without compromising flexibility.
On-device CNN inference for real-time computer vision applications can result in computational demands that far exceed the energy budgets of mobile devices.
%Aggressively optimized hardware accelerators can improve performance at the cost of programmability.
This paper proposes FixyNN, a co-designed hardware accelerator platform which splits a CNN model into two parts: a set of layers that are fixed in the hardware platform as a front-end fixed-weight feature extractor, and the remaining layers which become a back-end classifier running on a conventional programmable CNN accelerator.
The common front-end provides ubiquitous CNN features for all FixyNN models, while the back-end is programmable and specific to a given dataset.
% Image classification models for FixyNN are trained end-to-end via transfer learning, with the common feature extractor representing the transfered part, and the programmable part being learnt on the target dataset.
Image classification models for FixyNN are trained end-to-end via transfer learning, with front-end layers fixed for the shared feature extractor, and back-end layers fine-tuned for a specific task.
%This arrangement provides a significant improvement in throughput and energy efficiency, while maintaining flexibility among various computer vision tasks.
% Experimental results demonstrate FixyNN hardware can achieve very high energy efficiencies up to 
%26.6 TOPS/W (
% $4.81 \times$ better than iso-area programmable accelerator. % We need to be careful what we are reporting here
Over a suite of six datasets, we trained models via transfer learning with an accuracy loss of $\leq 1\%$, resulting in 
%up to 11.2 TOPS/W -- 
a FixyNN hardware platform with nearly $2 \times$ better energy efficiency than a conventional programmable CNN accelerator of the same silicon area (i.e. hardware cost).
%Highly aggressive hardware specialization in the FFE makes FixyNN a significant step forward towards closing the energy efficiency gap for on-device real-time CV.
%At the same time, by leveraging transfer learning concepts, we can exploit aggressively optimized specialized hardware without sacrificing generalization.
%

\end{abstract}

\section{Introduction}
\label{sec:intro}

Emerging applications such as augmented/mixed reality, autonomous drones and automotive driver assistance demand \textit{on-device} computer vision (CV) features, such as image classification, object detection/tracking, and semantic segmentation.
In support of these applications, we've seen a marked increase in accuracy on such CV tasks in recent years, following the displacement of traditional hand-crafted feature extractors by convolutional neural networks (CNNs)~\cite{energygap}.
%approaches have rapidly displaced traditional hand-crafted feature extractors, such as Haar~\cite{violajones} and HOG~\cite{hog}. 
%This shift in focus is motivated by a marked increase in accuracy on key CV tasks~\cite{vgg}.
However, CNNs pose a number of challenges for on-device inference due to a vast increase in the amount of computation and storage required~\cite{energygap}, which must be met by the hardware platform.
Unfortunately, mobile device hardware resources are heavily constrained in terms of both energy consumption and also the silicon area of the system-on-chip (SoC) inside the device.
%Mobile devices exhibit constraints in the energy and silicon area that can be allocated to CV tasks, which limits the adoption of CNNs at high resolution and frame-rate (e.g. 1080p at 30 FPS).
Therefore, a gap in energy efficiency exists between the demands of real-time CV CNNs, and the power constraints of mobile devices.
This gap is severely compounded at high image resolution and frame rate (e.g. 1080p at 30FPS).

\begin{figure}[t]
%\vspace{-10pt}
\centering
\includegraphics[width=0.95\textwidth]{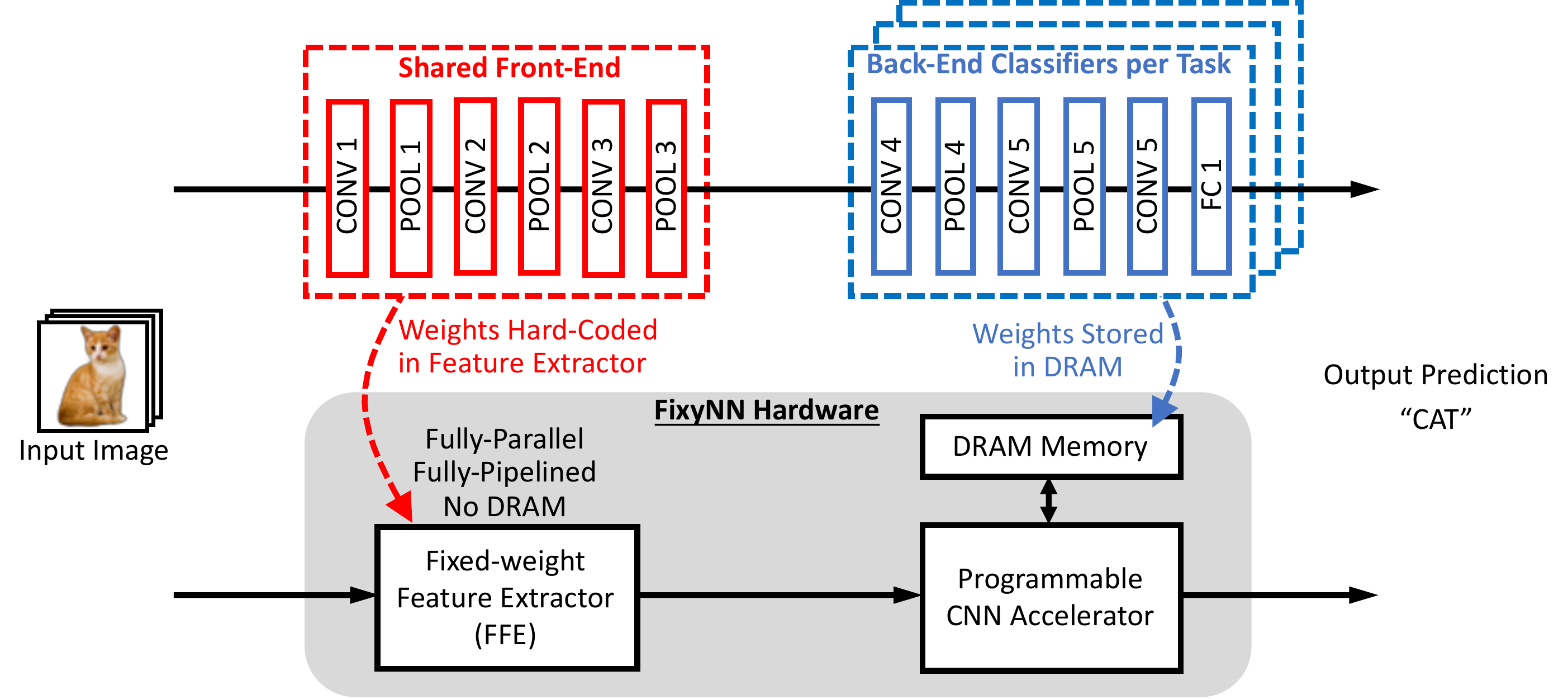}
\vspace{3pt}
\caption{FixyNN splits a deep CNN into two parts: a front-end fixed-weight feature extractor (FFE), and a back-end classifier.
The former is implemented as a heavily-optimized fixed-function hardware accelerator, whereas the latter is implemented on a canonical programmable accelerator.
Networks are trained via transfer learning to enable a single common FFE to be utilized for different tasks.
% \textbf{FixyNN} proposes to split a deep CNN into two parts, which are implemented in hardware using a (shared) fixed-weight feature extractor (FFE) hardware accelerator for the shared front-end and a canonical programmable accelerator for the task-specific back-end.
}
\vspace{-8pt}
\label{fig:concept}
\end{figure}

In this paper we describe \textbf{FixyNN}, which builds upon 
two key trends in on-device ML: more compact CNN architectures~\cite{MobileNetV1} and energy efficient CNN hardware accelerators~\cite{armml,nvdla}.
%both of these trends by means of a hardware/CNN co-design approach to efficient on-device CNN inference.
Section~\ref{sec:fixy} gives an overview of the FixyNN architecture, Section~\ref{sec:results} presents experimental results, and
Section~\ref{sec:conclusion} provides concluding remarks.

\section{FixyNN Overview}
\label{sec:fixy}

FixyNN is a CNN model architecture co-designed with the hardware platform and trained using transfer learning principles.
The general approach illustrated in Figure~\ref{fig:concept} is to divide a given model into a shared front-end \textit{fixed-weight feature extractor} (FFE) and a task-specific back-end classifier.
The FFE implements a fixed set of CNN layers that are common to all models, and is implemented as a heavily-optimized fixed-function hardware accelerator -- essentially an embodiment of ``do one thing and do it well''.
The FFE is fixed in hardware and used for all FixyNN models, and therefore the layers used are taken from a model trained on a large dataset such as ImageNet, to learn features that generalize well across a range of datasets.
The weights for the FFE are fixed in the hardware and do not require access to DRAM memory.
The back-end classifier
\footnote{The terms feature extractor and classifier are used very loosely here, merely to distinguish between the layers towards the front of the network, and the remaining layers up to the end.} 
is unique for each model, and is therefore implemented on a canonical programmable CNN hardware accelerator~\cite{nvdla,armml}, or could even be implemented using the mobile CPU or GPU.
The weights for the back-end classifier are stored in DRAM memory.
%Figure~\ref{fig:concept} illustrates this configuration.

% Our approach (Figure~\ref{fig:concept}) divides a CNN into two parts.
% The first part of the network implements a set of layers that are common for all CV tasks, essentially producing a set of universal low-level CNN features that are shared for multiple different tasks or datasets.
% The second part of the network provides a task-specific CNN back-end.
% These two CNN parts are then processed on different customized hardware.
% The front-end layers are implemented as a heavily optimized \textit{fixed-weight feature extractor (FFE)} hardware accelerator.
% The second part of the network is unique for each dataset, and hence must be implemented on a canonical programmable CNN hardware accelerator~\cite{nvdla,armod,armml}.

% TODO fully-parallel, fully-pipelined

In Section~\ref{sec:results}, we demonstrate significant throughput and energy efficiency advantage from the FixyNN hardware platform.
These gains are a result of diverting a significant portion of the computational load of a given CNN to the heavily-specialized FFE hardware accelerator.
The FFE can be heavily optimized because all the front-end CNN layers associated with the FFE are known and fixed at the time of designing the hardware platform.
Hence, we can aggressively exploit unstructured weight sparsity and other optimizations which typically offer little advantage to programmable hardware.
Usually, this kind of aggressive \textit{hardware specialization} has limited utility as the FFE is essentially a fixed-function accelerator and only implement a set of fixed-weight layers.
However, in the context of deep CNNs, it is well known that through \textit{transfer learning}~\cite{yosinski_nips14}, we are typically able to train new models that incorporate a set of front-end layers from a model trained on a different dataset.
Therefore, FixyNN explores an opportunity to aggressively exploit hardware specialization, without loosing the ability to generalize to a range of CV tasks.

\section{Experimental Results}
\label{sec:results}

%In this section, we present results from evaluating fixed feature extractors from the perspective of both the hardware latency and energy efficiency, as well as model generalization across domains.
%and also the model accuracy and generalization performance.
%We also explore the impact fixing different numbers of layers has on PPA.
%trade-offs in terms of the size of the fixed feature extractor and the impact this has on the silicon area, latency and energy.

Focusing on image classification problems, we evaluate FixyNN using experiments based on the MobileNetV1~\cite{MobileNetV1} CNN architecture, and focus on the compact MobileNet-0.25, which is well-suited to on-device inference applications.
%heavily constrained applications FixyNN is designed for~\footnote{FixyNN still provides benefits for larger networks, but at proportionally larger silicon area budgets.}.
%We first discuss the hardware performance of FixyNN, then explore the CNN generalization performance, and finally draw the two together with a discussion.
%We model the throughput and energy efficiency of FixyNN with MobileNet-0.25, with different numbers of  fixed front-end layers in the FFE and different sizes of programmable accelerator to perform task specific layers.
We evaluate both the hardware throughput and energy efficiency, as well as the model accuracy across a range of image classification datasets.

\begin{figure*}[t]
	%\vspace{-16pt}
	\centering
 	\subfloat[Throughput]{\label{fig:hybrid:tops}\includegraphics[width=0.45\textwidth]{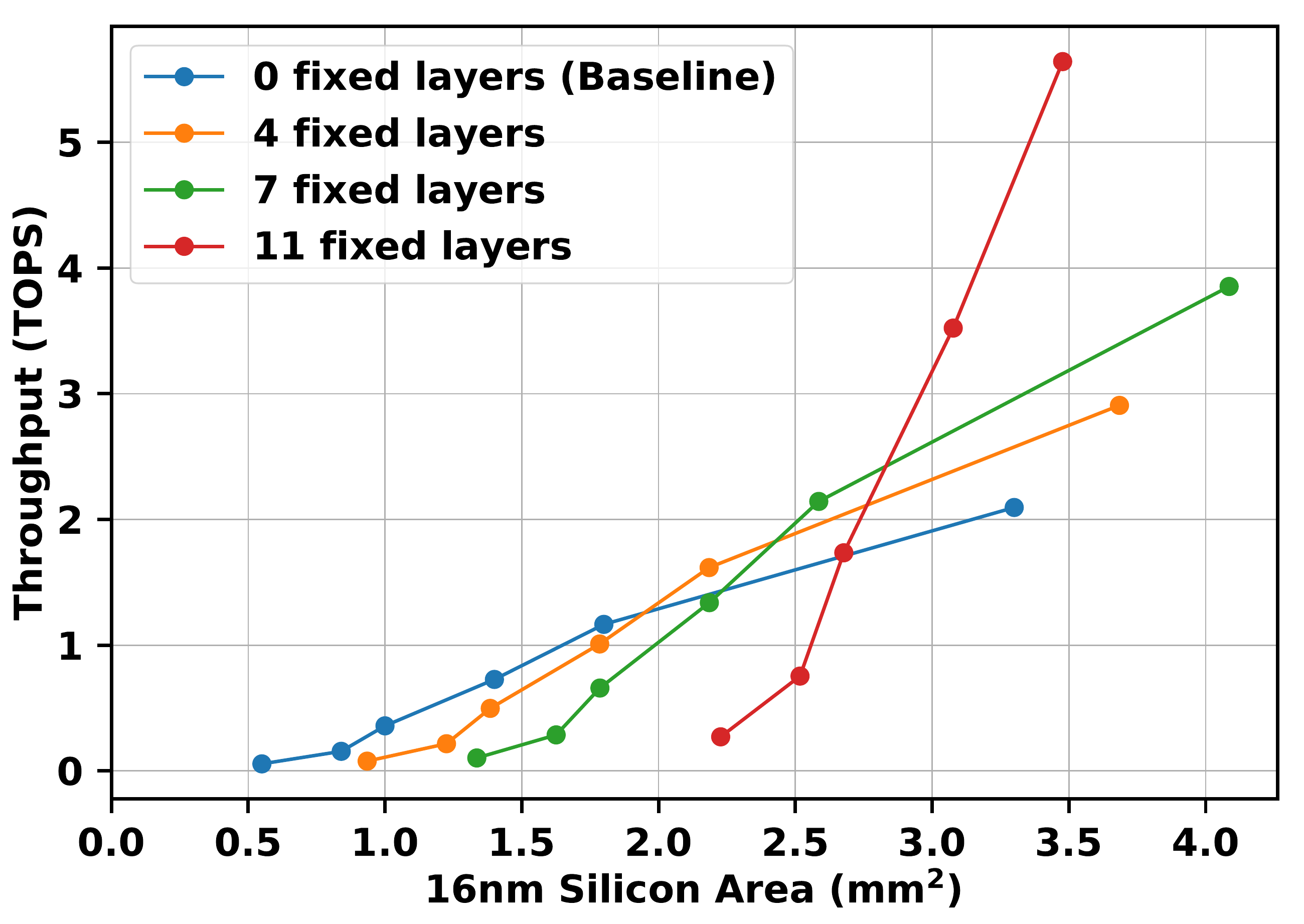}}
 	\subfloat[Energy Efficiency]{\label{fig:hybrid:topspw}\includegraphics[width=0.45\textwidth]{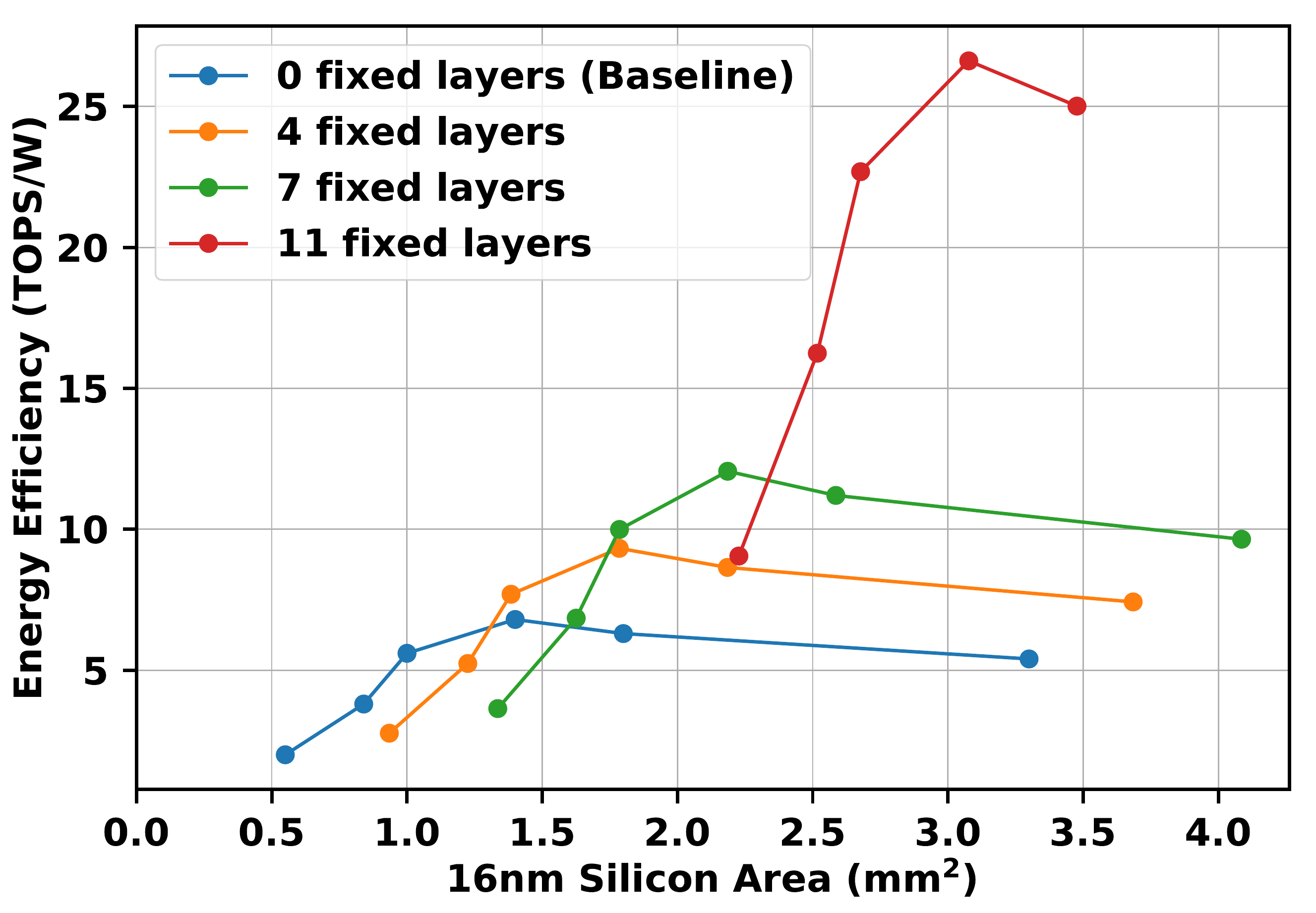}}
	\caption{
    Summary of FixyNN hardware results.
    Colors denote sweeping the number of layers fixed in the FFE.
    Points of the same color denote sweeping the hardware silicon area constraint in a 16nm silicon fabrication process.
% Each line corresponds to a single size feature extractor being used with different sized programmable accelerators.
}
	%\vspace{-16pt}
   \label{fig:hybrid}
\end{figure*}

\subsection{Hardware Evaluation}

%The FFE hardware accelerator has significantly higher throughput and energy-efficiency that a programmable hardware accelerator, such as Nvidia DLA (NVDLA)~\cite{nvdla}.
To implement the FFE hardware accelerator, we designed a tool called \textit{DeepFreeze} \cite{DeepFreeze}, which consumes a model described in a high-level framework such as TensorFlow, and generates a hardware description in the Verilog language.
DeepFreeze generates fully-parallel, fully-pipelined hardware, with optimizations for exploiting unstructured weight sparsity, and fine-grained quantization, as well as optimized storage which does not require DRAM access. 
%As well as optimizing the compute, the data storage in the FFE is implemented efficiently by pipelining only a small subset of the feature map data between network layers, rather than storing each feature map completely before starting to compute the next layer.
Since there are a variety of possible hardware configurations for FixyNN, with varying silicon area costs, we model the throughput
and energy efficiency 
for these configurations
\footnote{Throughput is reported in Terra-Operations Per Second (TOPS), and energy efficiency in TOPS/W.}
.
The FFE running the front-end model is modeled using hardware synthesis from a register-transfer level description, from which we can simulate clock frequency and power consumption.
The programmable back-end is modeled using previously published NVDLA data.
In all cases, we compare FixyNN with a baseline of a fully-programmable model running on NVDLA, which is the current state-of-the-art for mobile devices.
%We model performance and energy efficiency for different size FFEs from simulations of hardware description level (HDL) designs using Design Compiler and PrimeTime. % TODO: add specifics...
%We use published performance and power results of NVIDIA's DLA~\cite{nvdla} for modeling the programmable accelerator (however FixyNN can utilize any other programmable accelerator, taking advantage of any performance benefits they may provide).
%However, FixyNN will benefit from advancements in state-of-the-art of programmable accelerators since the programmable accelerator comprises the vast majority of latency and power in the system.

Figure~\ref{fig:hybrid} illustrates the throughput and energy efficiency trade-offs of fixing an increasing number of layers of a network in hardware at different hardware silicon area costs.
% Figure~\ref{fig:hybrid} illustrates the performance, power, and area (PPA) trade-offs of fixing different number of layers in the FFE, and pairing that with different sizes of programmable accelerator.
The results suggest that at areas greater than 2mm$^2$, it is beneficial in both performance and energy efficiency to invest area to fix some layers of a network with the FFE, rather than dedicating that area to a larger programmable accelerator (NVDLA).
However, it is inefficient to invest in an FFE at very low area budgets because even a small number of fixed layers consume a high percentage of the total area budget, causing a smaller programmable accelerator to bottleneck the system.

\begin{table*}[t]
\resizebox{\textwidth}{!}{%
\begin{tabular}{cc|cccccc|cc|cc}
\hline
\multicolumn{2}{c}{FixyNN configuration} & \multicolumn{6}{|c}{Accuracy (\%)} & \multicolumn{2}{|c}{Throughput} & \multicolumn{2}{|c}{Energy efficiency}   \\  Fixed layers & Fixed Ops (\%)  & CIFAR100   & CIFAR10    & SVHN   & Flwr   & Airc   & GTSR & TOPS & Relative & TOPS/W & Relative \\ \hline\hline
0  & 0.0  & 72.8 & 93.5 & 95.8 & 88.1 & 67.7 & 97.7 & 1.91 & 1.00$\times$ & 5.58 & 1.00$\times$ \\ 
4  & 27.1 & 72.5 & 93.3 & 95.7 & 88.3 & 66.7 & 97.8 & 2.32 & 1.21$\times$ & 7.98 & 1.43$\times$ \\
7  & 44.3 & 72.0 & 92.7 & 95.8 & 87.5 & 64.0 & 95.0 & 2.62 & 1.37$\times$ & 10.78 & 1.93$\times$ \\
11 & 77.0 & 71.1 & 91.7 & 94.6 & 86.9 & 56.7 & 89.2 & 3.18 & 1.66$\times$ & 25.86 & 4.63$\times$\\
14 & 97.0 & 68.5 & 85.3 & 91.0 & 82.8 & 41.9 & 59.3 & & & & \\%57.89 & 30.31$\times$ & 78.01 & 13.98$\times$ \\
\hline
\end{tabular}}
% \begin{tabular}{ccc|ccccccc}
% \hline
% \multicolumn{3}{c}{Model} & \multicolumn{7}{|c}{Accuracy on datasets (\%)} \\  Fixed layers & Adaptive BN & Fixed Ops (\%) & ImageNet  & CIFAR100   & CIFAR10    & SVHN   & Flwr   & Airc   & GTSR \\ \hline\hline
% 0 & N &0.0  & 49.8 & 72.8 & 93.5 & 95.8 & 88.1 & 67.7 & 97.7  \\ 
% 4 & Y & 27.1  & 49.8 & 72.5 & 93.3 & 95.7 & 88.3 & 66.7 & 97.8  \\
% 7 & Y & 44.3  & 49.8 & 72.0 & 92.7 & 95.8 & 87.5 & 64.0 & 95.0  \\
% 7 & N & 46.6  & 49.8 & 69.4 & 91.7 & 94.7 & 85.2 & 63.2 & 93.5  \\ 
% 11     & Y & 77.0  & 49.8 & 71.1 & 91.7 & 94.6 & 86.9 & 56.7 & 89.2  \\
% 14      &Y & 97.0 & 49.8 & 68.5 & 85.3 & 91.0   & 82.8 & 41.9 & 59.3  \\
% 14             & N & 100.0 & 49.8 & 54.5 & 77.0 & 48.0 & 77.8 & 30.5 & 46.1  \\   
% \hline
% \end{tabular}}
\caption{Results for MobileNet-$0.25$ with a fixed-weight feature extractor.
The model is trained on ImageNet and the FFE layers are transferred to six different vision datasets.
All BN parameters in the FFE are fine-tuned on the target dataset, with all other parameters fixed.
Hardware results for performance and energy efficiency assume a 3.0mm$^2$ area budget.}
\label{table:ml_results_1}
\end{table*}

\subsection{ML Evaluation}

FixyNN utilizes transfer learning concepts to train models for various datasets that take advantage of the single, shared FFE front-end.
Similar to \cite{yosinski_nips14}, each model is pretrained on ImageNet and the back-end layers are fine-tuned on each target dataset.
However, we do enable training of batch norm (BN) parameters across all layers (i.e. including the layers in the FFE), which we found to dramatically improve performance with a negligible area and power increase in the FFE.
%This requires programmability of a small number of parameters in the FFE.
%In terms of the ML performance, 
Table~\ref{table:ml_results_1} summarizes the accuracies for our transfer learning experiments with MobileNet-$0.25$, where the model is pretrained on ImageNet to a top-1 accuracy of $49.8\%$ and then transferred to CIFAR-10, CIFAR-100, Street View House Numbers (SVHN), Flowers102 (Flwr), FGVC-Aircraft (Airc), and German Traffic Sign Recognition benchmark (GTSR). 
%FC layers are omitted from hardware evaluation since they differ in size for each dataset and they comprise a small percentage of total ops in the network.
%This set comprises a range of different CV tasks.
The first row (our baseline) shows the performance of the model fully fine-tuned to the new tasks.
As more layers are fixed in the network, a bigger FFE is used.
At $3\mathrm{mm}^2$ area budget, hardware performance can significantly benefit from the bigger FFE.
However, deeper layers are associated with more task specific features. 
Fixing more layers in the FFE generally results in loss of model accuracy.
For datasets CIFAR-10, CIFAR-100, SVHN and Flwr, $77\%$ of the network can be fixed with less than $2\%$ loss in model accuracy. 
For datasets Airc and GTSR, similar accuracy performance relative to the baseline requires fixing a smaller percentage of the network in the FFE (between $27\%$ and $44\%$).

\subsection{Discussion}

Our experiments suggest that dedicating some percentage of the hardware platform to a fixed-weight feature extractor provides significant performance and power benefits when the area devoted to CNN vision tasks is greater than 2mm$^2$.
Fixing more layers of a network provides better hardware performance by diverting computational load from the programmable accelerator to the highly-efficient FFE.
However, as more layers are fixed, the task of training a new network incorporating the FFE on a different dataset becomes more challenging, and tends to incur an accuracy loss.
%Table~\ref{table:ml_results_1} indicates that a FFE may generalize to some tasks better than others.
%Therefore, 
%These insights indicate that it is imperative to balance the requirements of throughput/energy-efficiency and accuracy across a variety of tasks.
Concretely, we found that fixing 7 layers of MobileNet-0.25 in hardware, we have shown that FixyNN achieves 1.37$\times$ and 1.93$\times$ better performance and energy efficiency, respectively, with $\leq 1\%$ accuracy loss, compared to a traditional programmable accelerator.
Since two tasks (Airc, GTSR) incur a potentially unacceptable accuracy loss with 7 fixed layers, we propose to modify FixyNN to provide access to output activations from an earlier layer in the FFE, such that a model does not have to use all the fixed layers in the FFE.
With this modification, models for Airc and GTSR can use just 4 layers of the 7 layer FFE, resulting in a 1.04$\times$ and 1.49$\times$ improvement in performance and energy efficiency for these datasets, while still achieving the 1.37$\times$ and 1.93$\times$ improvement for the remainder.
All tasks incur $\leq$1\% absolute accuracy loss in this configuration.
%Hence, the advantages we see with FixyNN are also likely to increase with any future advancements in training techniques around transfer learning.

\section{Conclusion}
\label{sec:conclusion}

In this paper, we evaluate FixyNN, which proposes to split a CNN model into two components: a common front-end which we compute in a heavily optimized fixed-weight feature extractor (FFE), and a programmable back-end implemented on a conventional programmable hardware accelerator.
This combination allows us to take advantage of aggressive hardware specialization for the front-end, but retain generalization to a range of datasets by training the programmable back-end using transfer-learning principles.
Experimental results show that FixyNN provides nearly $2\times$ improvement in on-device CNN energy efficiency, with an accuracy degradation of $\leq 1\%$.
This is a significant step forward towards the goal of real-time, on-device CNN inference.
While fixing even more layers would result in higher hardware throughput and energy efficiency, it can also lead to prediction accuracy degradation.
Therefore, balancing the number of fixed layers is crucial in FixyNN hardware platforms.
Finally, we note that future research progress with transfer learning research is likely to further strengthen the case for hardware specialization for CNNs.

\bibliography{nns}
\bibliographystyle{unsrt}

\end{document}